\documentclass[11pt,letterpaper]{article}
\usepackage[letterpaper]{geometry}
\usepackage{acl2012}
\usepackage{times}
\usepackage{latexsym}
\usepackage{amsmath}
\usepackage{multirow}
\usepackage{url}
\usepackage{graphicx}
\usepackage{color}
\usepackage{helvet}
\usepackage{courier}
\usepackage{amssymb}
\usepackage{amsopn}
\usepackage{stfloats}
\usepackage[american]{babel}

\makeatletter
\newcommand{\@BIBLABEL}{\@emptybiblabel}
\newcommand{\@emptybiblabel}[1]{}
\makeatother
\usepackage[hidelinks]{hyperref}

\DeclareMathOperator{\Attention}{Attention}
\DeclareMathOperator{\Mean}{Mean}
\DeclareMathOperator{\LSTM}{LSTM}
\DeclareMathOperator{\Softmax}{Softmax}
\DeclareMathOperator{\Prob}{P}
\DeclareMathOperator{\linear}{linear}

\newcommand{\pheadB}[1] {\vspace{1mm}\noindent\textbf{#1}}

\title{Video Captioning with Multi-Faceted Attention}

\author{Xiang Long\\
  IIIS, Tsinghua University \\
  {longx13@mails.tisnghua.edu.cn} \\\And
  Chuang Gan \\
  IIIS, Tsinghua University \\
  {ganchuang1990@gmail.com} \\\And
  Gerard de Melo\\
  Rutgers University \\
  {gdm@demelo.org} \\}

\date{}

\begin{document}
\maketitle
\begin{abstract}
    Recently, video captioning has been attracting an increasing amount of interest, due to its potential for improving accessibility and information retrieval. While existing methods rely on different kinds of visual features and model structures, they do not fully exploit relevant semantic information. We present an extensible approach to jointly leverage several sorts of visual features and semantic attributes. Our novel architecture builds on LSTMs for sentence generation, with several attention layers and two multimodal layers. The attention mechanism learns to automatically select the most salient visual features or semantic attributes, and the multimodal layer yields overall representations for the input and outputs of the sentence generation component. Experimental results on the challenging MSVD and MSR-VTT datasets show that our framework outperforms the state-of-the-art approaches, while ground truth based semantic attributes are able to further elevate the output quality to a near-human level.
\end{abstract}

\section{Introduction}
    The task of automatically generating captions for videos has been receiving an increasing amount of attention.

    On YouTube, for example, every single minute, hundreds of hours of video content are uploaded. There is no way a person could sit and watch these overwhelming amounts of videos, so
    new techniques to search and quickly understand them are highly sought. Generating captions, i.e., short natural language descriptions, for videos is an important technique to address this challenge, while also greatly improving their accessibility for blind and visually impaired users.

    Video captioning has been studied for a long time and remains challenging, given the difficulties of video interpretation, natural language generation, and the interplay between them. Understanding a video hinges on our ability to make sense of video frames and of the relationships between consecutive frames.

    The output needs to be grammatically correct sequence of words. Different parts of the output caption may pertain to different parts of the video. In previous work, 3D ConvNets \cite{Du2014C3D} have been proposed to capture motion information in short videos, while LSTMs \cite{Hochreiter1997Long} can be used to generate natural language, and a variety of different visual attention models \cite{Yao2015Describing,Pan2015Hierarchical,Yu2015Video} have been deployed, attempting to capture the relationship between caption words and the video content.

    These methods, however, only make use of visual information from the video, often with unsatisfactory results. In many real-world settings, we can easily obtain additional information related to the video. Apart from sound, there may also be a title, user-supplied tags, categories, and other metadata. Both visual video features as well as attributes such as tags can be imperfect and incomplete.
    However, by jointly considering all available signals, we may obtain complementary information that aids in generating better captions. Humans, too, often benefit from additional context information when trying to understand what a video is portraying.

    Incorporating these additional signals is not just a matter of adding additional features. While generating the sequence of words in the caption, we need to be able to flexibly attend to the relevant frames over time, the relevant parts within a given frame, and relevant additional signals to the extent that they pertain to a particular output word.

\begin{figure}[t]
        \centering
        \includegraphics[width=0.48\textwidth]{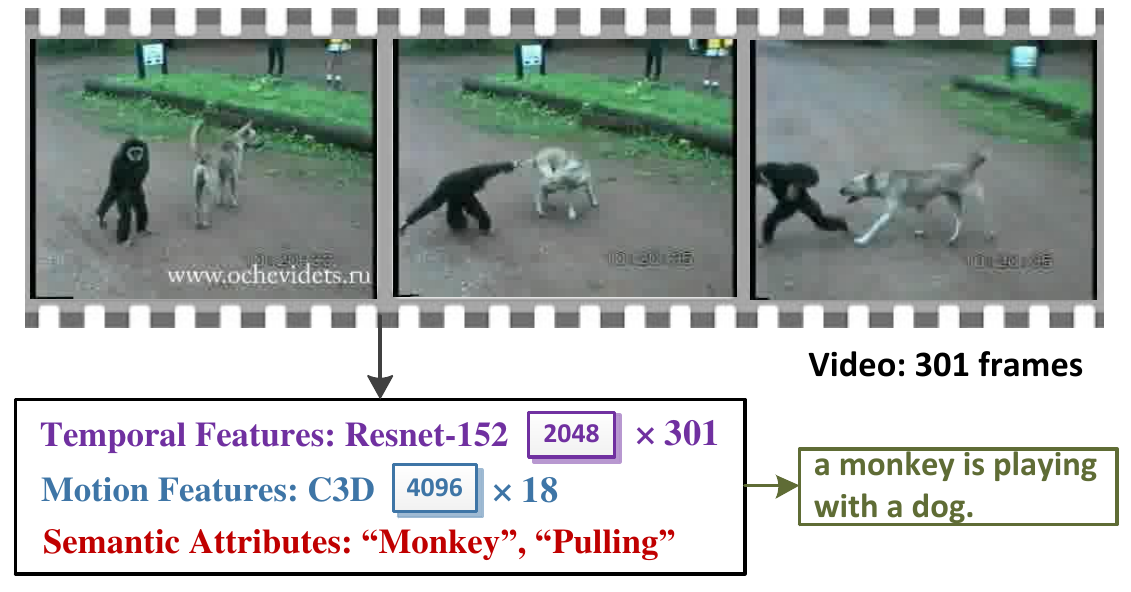}\\
        \caption{Example video with extracted visual features, semantic attribute, and the generated caption as output. }
        \label{fig:feature}
    \end{figure}

	Based on these considerations, we propose a novel multi-faceted attention architecture that jointly considers multiple heterogeneous forms of inputs. This model is flexibly attends to temporal information, motion features, and semantic attributes for every channel. An example of this is given in Figure~\ref{fig:feature}. Each part of the attention model is an independent branch and it is straightforward to incorporate additional branches for further kinds of features, making our model highly extensible. We present a series of experiments that highlight the contribution of attributes to yield state-of-the-art results on standard datasets.

\section{Related Work}

    \noindent \textbf{Machine Translation}. Some of the first widely noted successes of deep sequence-to-sequence learning models were for the task of machine translation \cite{Cho2014Learning,Cho2014On,Sutskever2014Sequence,Kalchbrenner2013Recurrent,Li2015A,Lin2015Hierarchical}. In several respects, this is actually a similar task to video caption generation, just with a rather different input modality. What they share in common is that both require bridging different representations, and that often an encoder-decoder paradigm is used with a Recurrent Neural Network (RNN) decoder to generate sentences in the target language. Many techniques for video captioning are inspired by neural machine translation ones, including soft attention mechanisms to focus on different parts of the input when generating the target sentence word by word \cite{Bahdanau2015Neural}.

    \pheadB{Image Captioning}. Image captioning can be regarded as a greatly simplified case of video captioning, with videos consisting of just a single frame. Recurrent architectures are often used here as well \cite{Karpathy2014Large,Kiros2014Unifying,Chen2015Learning,Mao2015Deep,Vinyals2014Show}.
    Spatial attention mechanisms allow for focusing on different areas of an image \cite{Xu2015Show}. Recently, image captioning incorporating semantic concepts have achieved inspiring results. A semantic attention approach has been proposed \cite{You2016Image} to selectively attend to semantic concept proposals and fuse them into hidden states and outputs of RNNs, but
    their model is difficult to extend for multiple channels.
    Overall, none of these methods for image captioning need to account for temporal and motion aspects.

    \pheadB{Video captioning}. For video captioning, many works utilize a recurrent neural architecture to generate video descriptions, conditioned on either an average-pooling \cite{Venugopalan2015Translating} or recurrent encoding \cite{Xu2015A,Donahue2015Long,Venugopalan2015Sequence,Venugopalan2016Improving} of frame-level features, or on a dynamic linear combination of context vectors obtained via temporal attention \cite{Yao2015Describing}. Recently,  hierarchical recurrent neural encoders (HRNE) with attention mechanism have been proposed to encode video \cite{Pan2015Hierarchical}. A recent paper \cite{Yu2015Video} additionally exploits several kinds of visual attention and relies on a multimodal layer to combine them. In our work, we present a novel attention model with more effective multimodal layers that jointly models multiple heterogeneous signals, including semantic attributes, and experimentally show the benefits of this approach over previous work.

\section{The Proposed Approach}
\begin{figure*}[t]
        \centering
        \includegraphics[width=1.0\textwidth]{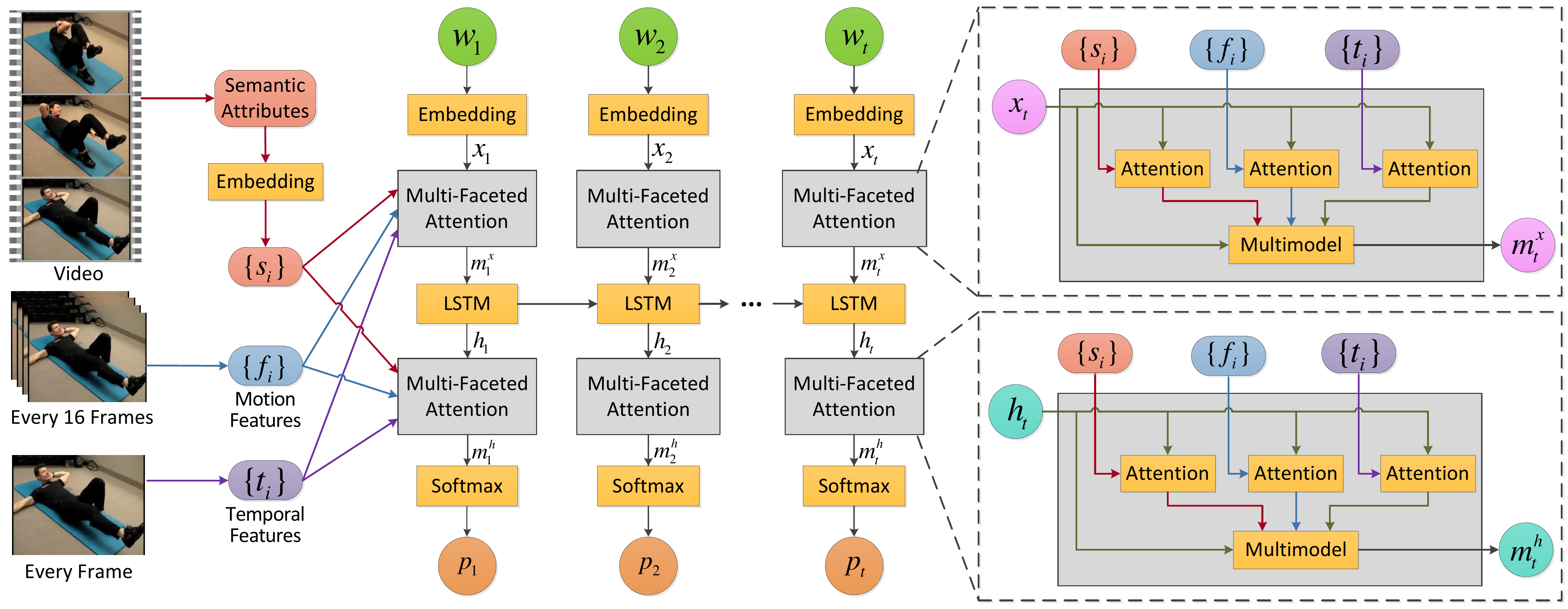}\\
        \caption{Model architecture. Temporal, motion, and semantic features are weighted via an attention mechanism and aggregated, together with an input word, by a multimodal layer. Its output updates the LSTM's hidden state. A similar attention mechanism then determines the next output word via a softmax layer.}
        \label{fig:model}
    \end{figure*}

    In this section, we describe our approach for combining multiple forms of attention for video captioning. Figure \ref{fig:model} illustrates the architecture of our model. The core of our model is a sentence generator based on generator is a simple Long Short Term Memory (LSTM) units \cite{Hochreiter1997Long}.
    Instead of a traditional sentence generator, which directly receives a previous word and selects the next word, our model relies on several attention layers to selectively focus on important parts of temporal, motion, and semantic features. The output words are generated via a softmax reading from a multimodal layer \cite{Mao2015Deep}, which integrates information from the different attention layers. An additional multimodal layer integrates information before the input reaches the sentence generator to enable better hidden representations in the LSTM. We first briefly review the basic LSTM, and then describe our model in detail, including our novel multi-faceted attention mechanism to consider temporal, motion, and semantic attribute perspectives.

    \subsection{Long Short Term Memory Networks}
        A Recurrent Neural Network (RNN) \cite{Elman1990Finding} is a neural network adding extra feedback connections to feed-forward networks, so as to be able to work with sequences. The network is updated not only based on the input but also based on the previous hidden state. RNNs can compute the hidden states $(h_1,h_2,\dots,h_m)$ given an input sequence $(x_1,x_2,\dots,x_m)$ based on recurrence of the following form:
        \begin{equation}
            h_t=\phi(W_h x_t+U_h h_{t-1}+b_h),
        \end{equation}
        where weight matrices $W$, $U$ and bias $b$ are parameters to be learned and $\phi(\cdot )$ is an element-wise activation function.

        RNNs trained via unfolding have proven inferior at capturing long-term temporal information. LSTM units were introduced to avoid these challenges. LSTMs not only compute the hidden states but also maintains a cell state to account for relevant signals that have been observed. They have the ability to remove or add information to the cell state, modulated by gates.

        Given an input sequence $(x_1,x_2,...,x_m)$, an LSTM unit computes the hidden state $(h_1,h_2,...,h_m)$ and cell states $(c_1,c_2,...,c_m)$ via repeated application of the following equations:
        \begin{align}
            i_t &=\sigma(W_i x_t+U_i h_{t-1}+b_i)\\
            f_t &=\sigma(W_f x_t+U_f h_{t-1}+b_f)\\
            o_t &=\sigma(W_o x_t+U_o h_{t-1}+b_o)\\
            g_t &=\phi(W_g x_t+U_g h_{t-1}+b_g)\\
            c_t &= f_t \odot c_{t-1} + i_t \odot g_t\\
            h_t &= o_t \odot c_t,
        \end{align}
        where $\sigma(\cdot)$ is the sigmoid function and $\odot$ denotes the element-wise multiplication of two vectors. For convenience, we denote the computations of the LSTM at each time step $t$ as $h_t$, $c_t = \LSTM(x_t, h_{t-1}$, $c_{t-1})$.

    \subsection{Input Representations}

        When training a video captioning model, as a first step, we need to extract feature vectors that serve as inputs to the LSTM. For visual features, we can extract one feature vector per frame, leading to a series of what we call \emph{temporal features}. We can also extract another form of feature vector from several consecutive frames, which we call \emph{motion features}. Additionally, we could also extract other forms of visual features, such as features from an area of a frame, the same area of consecutive frames, etc. In this paper, we only consider temporal features, denoted by $\{ v_i \}$, and motion features, denoted by $\{ f_i \}$, which are commonly used in video captioning.

        For semantic features, we need to extract a set of related attributes denoted by $\{ a_i \}$. These can be based on title, tags, etc., if available. Alternatively, we can also rely on techniques to extract or predict attributes that are not directly given. In particular, because we have captions for the videos in the training set, we can train different models to predict caption-related semantic features for videos in the validation and test sets.
        As the choice of semantic features is not the core contribution, we describe our specific experimental setups in Section~\ref{sec:experiments}.

        After determining a set of attributes for each video, each attribute $a_i$ in any video in the entire dataset corresponds to an entry in the vocabulary and each word $w_i$ in any caption of the training set also corresponds to an entry in the vocabulary.

        An embedding matrix $E$ is used to represent both words and semantic attributes and we denote by $E[w]$ an embedding vector of a given $w$. Thus, we obtain attribute embedding vectors $\{ s_i \}$ and input word embedding vectors as:
        \begin{align}
            s_i &= E[a_i]\\
            x_t &= E[w_t]
        \end{align}

    \subsection{Multi-Faceted Attention}

        We do not directly feed $x_t$ to the LSTM. Instead, we first apply our multi-faceted attention model to $x_t$.

        Assuming that we have a series of multimodal feature vectors for a given video, we generate a caption word by word. At each step, we need to select relevant information from these feature vectors, which we from now on refer to as \emph{context vectors} $\{ c_1,c_2,...,c_n\}$. Due to the variability of the length of videos, it is challenging to directly input all these vectors to the model at every time step. A simple strategy is to compute the average of the context vectors and input this average vector to each time step of the model.
                \begin{equation}
                y_t = \frac{1}{m} \sum_{i=1}^m c_i
                \end{equation}
                However, this strategy collapses all available information into a single vector, neglecting the inherent structure, which captures the temporal progression, among other things. Thus, this sort of folding leads to a significant loss of information. Instead, we wish to focus on the most salient parts of the features at every time step. Instead of a naive averaging of the context vectors $\{ c_1,c_2,\dots,c_n\}$, a soft attention model calculates weights $\alpha_i^t$ for each $c_i$, conditioning on the input vector $x_t$ at each time step $t$. For this, we first compute basic attention scores $e_i^t$ and then feed these through
                a sequential softmax layer to obtain a set of attention weights $\{ \alpha_1^t,\alpha_2^t,\dots,\alpha_n^t\}$ that quantify the relevance of $\{ c_1,c_2,\dots,c_n\}$ for $x_t$.
                \begin{align}
                e_i^t &= x_t^T U c_i\\
                \alpha_i^t &= \frac{\exp(e_i^t)}{\sum_{j=1}^n \exp(e_j^t)}\\
                y_t &= \sum_{i=1}^m \alpha_i^t c_i
                \end{align}
                We obtain the corresponding output vectors $y_t$ as weighted averages.

                This soft attention model, strictly speaking, converts an entire input sequence $(x_1,x_2,\dots,x_m)$ to an entire output sequence $(y_1,y_2,\dots,y_m)$ based on all context vectors $\{ c_1,c_2,\dots,c_n\}$. For convenience, we denote the attention model outputs at a given time step $t$ as $y_t = \Attention(x_t, \{ c_i\})$.

        In particular, the attention model is applied to the temporal features $\{ v_i \}$, motion features $\{ f_i \}$ and semantic features $\{ s_i \}$:
        \begin{align}
            s_t^{x} &= \Attention(x_t,\{ s_i \})\\
            v_t^{x} &= \Attention(x_t,\{ v_i \})\\
            f_t^{x} &= \Attention(x_t,\{ f_i \})
        \end{align}

        We then obtain the input to the LSTM $m_t^x$ via a multimodal layer
        \begin{equation}
            m_t^x =\phi \left. (W^{x} \left. [x_t,s_t^{x},w_v^x \odot v_t^{x},w_f^x \odot f_t^{x}] \right. +b^{m,x}) \right.
        \end{equation}

        Here, $w_v^x$ and $w_f^x$ facilitate capturing the relative importance of each dimension of the temporal and motion feature space \cite{You2016Image}. We apply dropout \cite{Srivastava2014Dropout} to this multimodal layer to reduce overfitting.

        Subsequently, we can obtain $h_t$ via the LSTM. At the first time step, the mean values of the features are used to initialize the LSTM states to yield a general overview of the video:
        \begin{equation}
            m_0^x=W^{i}[\Mean(\{ s_i \}),\Mean(\{ v_i \}),\Mean(\{ f_i \})]
        \end{equation}
        \begin{equation}
            h_0,c_0= \LSTM(m_0^x,0,0)
        \end{equation}
        \begin{equation}
            h_t,c_t= \LSTM(m_t^x,h_{t-1},c_{t-1})
        \end{equation}
        where $\Mean(\cdot )$ denotes mean pooling of the given feature set.

        We also apply the attention model to hidden states $h_t$, and use a multimodal layer to concatenate outputs of the attention model and map it into a feature space that has exactly the same dimensionality as the word embeddings. This multimodal layer is followed by a softmax layer with a dimensionality equal to the size of the vocabulary. The projection matrix from the multimodal layer to the softmax layer is set to be the transpose of the word embedding matrix:
        \begin{align}
            s_t^{h} &= \Attention(h_t,\{ s_i \})\\
            v_t^{h} &= \Attention(h_t,\{ v_i \})\\
            f_t^{h} &= \Attention(h_t,\{ f_i \})
        \end{align}
        \begin{equation}
            m_t^h=\phi \left. (W^{h} \left. [h_t,s_t^{h},w_v^h \odot v_t^{h},w_f^h \odot f_t^{h}] \right. +b^{m,h}) \right.
        \end{equation}
        \begin{equation}
            p_t = \Softmax(E^T m_t^h)
        \end{equation}
        where $\Softmax(\cdot)$ denotes a sequential softmax.

        By using two multimodel layers, we combine six attention layers with the core LSTM. This model is highly extensible since we can easily add extra branches for additional features.

    \subsection{Training and Generation}
        We can interpret the output of the softmax layer $p_t$ as a probability distribution over words:
        \begin{equation}
            \Prob(w_{t+1}|w_{1:t},V,S,\Theta)
        \end{equation}
        where $V$ denotes the corresponding video, $S$ denotes semantic attributes and $\Theta$ denotes model parameters. The overall loss function is defined as the negative logarithm of the likelihood and our goal is to learn all parameters $\Theta$ in our modal by minimizing the loss function over the entire training set:
        \begin{equation}
            \min_{\Theta} \quad -\frac{1}{N} \sum_{i=1}^N \sum_{t=1}^{T_i} \Prob(w_{t+1}^i|w_{1:t}^i,V^i,S^i,\Theta)
        \end{equation}
        where $N$ is the total number of captions in the training set, and $T_i$ is the number of words in caption $i$. During the training phase, we add a begin-of-sentence tag $\langle$BOS$\rangle$ to the start of the sentence and an end-of-sentence tag $\langle$EOS$\rangle$ to the end of sentence. We use Stochastic Gradient Descent to find the optimum with the gradient computed via Backpropagation Through Time (BPTT) \cite{Werbos1990Backpropagation}. Training continues until the METEOR evaluation score on the validation set stops increasing, and we optimize the hyperparameters using random search to maximize METEOR on the validation set, following previous studies that found that METEOR is more consistent with human judgments than BLEU or ROUGE \cite{Vedantam2015CIDEr}.

        After the parameters are learned, during the testing phase, we also have temporal and motion features extracted from the video as well as semantic attributes, which were either already given or are predicted using a model trained on the training set. Given a previous word, we can calculate the probability distribution of the next word $p_t$ using the model described above. Thus, we can generate captions starting from the special symbol $\langle$BOS$\rangle$ with Beam Search \cite{Yu2015Video}.

\begin{table*}[t]
        \centering
        \resizebox{0.9\textwidth}{!}{
        \begin{tabular}{l|*{4}c|c|c}
        \hline
        Model & BLEU-1 & BLEU-2 & BLEU-3 & BLEU-4 & METEOR  & CIDEr \\
        \hline
        \hline
        LSTM-YT \cite{Venugopalan2015Translating} & - & - & - & 0.333 & 0.291 & - \\
        S2VT \cite{Venugopalan2015Sequence} & - & - & - & - & 0.298 & - \\
        TA \cite{Yao2015Describing}  & 0.800 & 0.647 & 0.526 & 0.419 & 0.296 & 0.517\\
        TA$^*$  & 0.811 & 0.655 & 0.541 & 0.422 & 0.304 & 0.524 \\
        LSTM-E \cite{Pan2016Jointly} & 0.788 & 0.660 & 0.554 & 0.453 & 0.310 & - \\
        HRNE-A \cite{Pan2015Hierarchical} & 0.792 & 0.663 & 0.551 & 0.438 & 0.331 & - \\
        h-RNN \cite{Yu2015Video} & 0.815 & 0.704 & 0.604 & 0.499 & 0.326 & 0.658 \\
        h-RNN$^*$ & 0.824 & 0.711 & 0.610 & 0.504 & 0.329 & 0.675 \\
        \hline
        T (Ours) & 0.813 & 0.698 & 0.605 & 0.504 & 0.322 & \textbf{0.698} \\
        M (Ours) & 0.807 & 0.682 & 0.583 & 0.473 & 0.308 & 0.656 \\
        TM (Ours) & \textbf{0.826} & \textbf{0.717} & \textbf{0.619} & \textbf{0.508} & \textbf{0.332} & 0.694  \\
        \hline
        \end{tabular}
        }
        \caption{Results of models using visual features only, on MSVD, where
        	(-) indicates unknown scores.}
        \label{tab:results1}
    \end{table*}

   \begin{table*}[t]
        \centering
        \resizebox{0.8\textwidth}{!}{
        \begin{tabular}{l|*{4}c|c|c|c}
        \hline
        Model & BLEU-1 & BLEU-2 & BLEU-3 & BLEU-4 & METEOR  & CIDEr & ROUGE-L \\
        \hline
        \hline
        TM & 0.826 & 0.717 & 0.619 & 0.508 & 0.332 & \textbf{0.694} & 0.702  \\
        TM-P-NN & 0.809 & 0.696 & 0.604 & 0.508 & 0.324 & 0.693 & 0.682 \\
        TM-P-SVM & 0.814 & 0.719 & 0.620 & 0.512 & 0.330 & 0.679 & 0.703 \\
        TM-P-HRNE & \textbf{0.829} & \textbf{0.720} & \textbf{0.627} & \textbf{0.528} & \textbf{0.334} & 0.689 & \textbf{0.705} \\
        \hline
        TM-HQ-S & 0.863 & 0.807 & 0.740 & 0.633 & 0.367 & 0.899 & 0.747 \\
        TM-HQ-V & 0.877 & 0.788 & 0.710 & 0.639 & 0.371 & 1.075 & 0.749 \\
        TM-HQ-SV & \textbf{0.918} & \textbf{0.872} & \textbf{0.825} & \textbf{0.764} & \textbf{0.429} & \textbf{1.393} & \textbf{0.814} \\
        \hline
        Human & 0.891 & 0.776 & 0.681 & 0.583 & 0.436 & 1.322 & 0.761 \\
        \hline
        \end{tabular}
        }
        \caption{Results of models combining visual and semantic attention on MSVD.
        	}
        \label{tab:results2}
    \end{table*}

 \begin{table*}[t]
        \centering
        \resizebox{0.65\textwidth}{!}{
        \begin{tabular}{l|c|c|c|c}
        \hline
        Model & BLEU-4 & METEOR  & CIDEr & ROUGE-L\\
        \hline
        \hline
        Rank: 1, Team: v2t\_navigator& \textbf{0.408} & \textbf{0.282} & 0.448 & \textbf{0.609} \\
        Rank: 2, Team: Aalto	& 0.398 & 0.269 & \textbf{0.457} & 0.598 \\
        Rank: 3, Team: VideoLAB	 & 0.391 & 0.277 & 0.441 & 0.606 \\
        \hline
        T  & 0.367  & 0.257  & 0.400  & 0.581 \\
        M  & 0.361  & 0.253  & 0.392  & 0.577  \\
        TM & 0.386  & 0.265  & 0.439  & 0.596  \\
        TM-P-NN  & 0.370  & 0.254  & 0.401  & 0.579 \\
        TM-P-SVM & 0.376  & 0.259  & 0.412  & 0.585  \\
        TM-P-HRNE  & \textbf{0.392}  & \textbf{0.269}  & \textbf{0.446} & \textbf{0.601} \\
        \hline
        TM-HQ-S & 0.421  & 0.286  & 0.478  & 0.610  \\
        TM-HQ-V & 0.429  & 0.289  & 0.481  & 0.613  \\
        TM-HQ-SV &\textbf{0.451}  & \textbf{0.292}  & \textbf{0.503} &  \textbf{0.625} \\
        \hline
        Human & 0.343  & 0.295  & 0.501 &0.560 \\
        \hline
        \end{tabular}
        }
        \caption{Results on MSR-VTT.}
        \label{tab:results3}
    \end{table*}

    \begin{figure*}[t]
        \centering
        \includegraphics[width=0.88\textwidth]{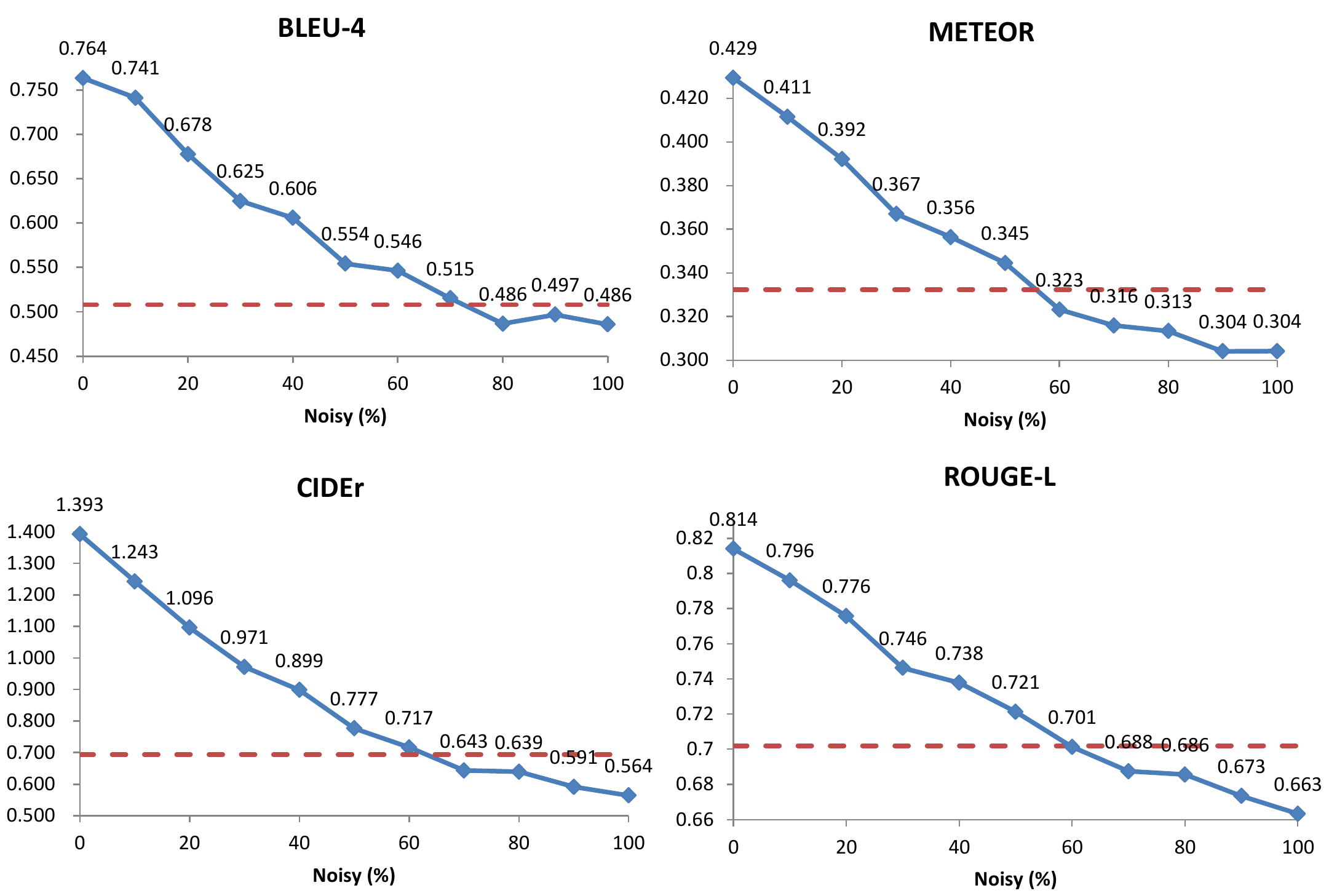}\\
        \caption{Results of adding noise to high-quality semantic attributes of MSVD. The \textcolor[rgb]{0,0.44,0.75}{blue} solids are results of adding noise. The \textcolor[rgb]{0.75,0,0}{red} dashes are corresponding results of the TM model.}
        \label{fig:result}
    \end{figure*}

    \begin{figure*}[t]
        \centering
        \includegraphics[width=0.9\textwidth]{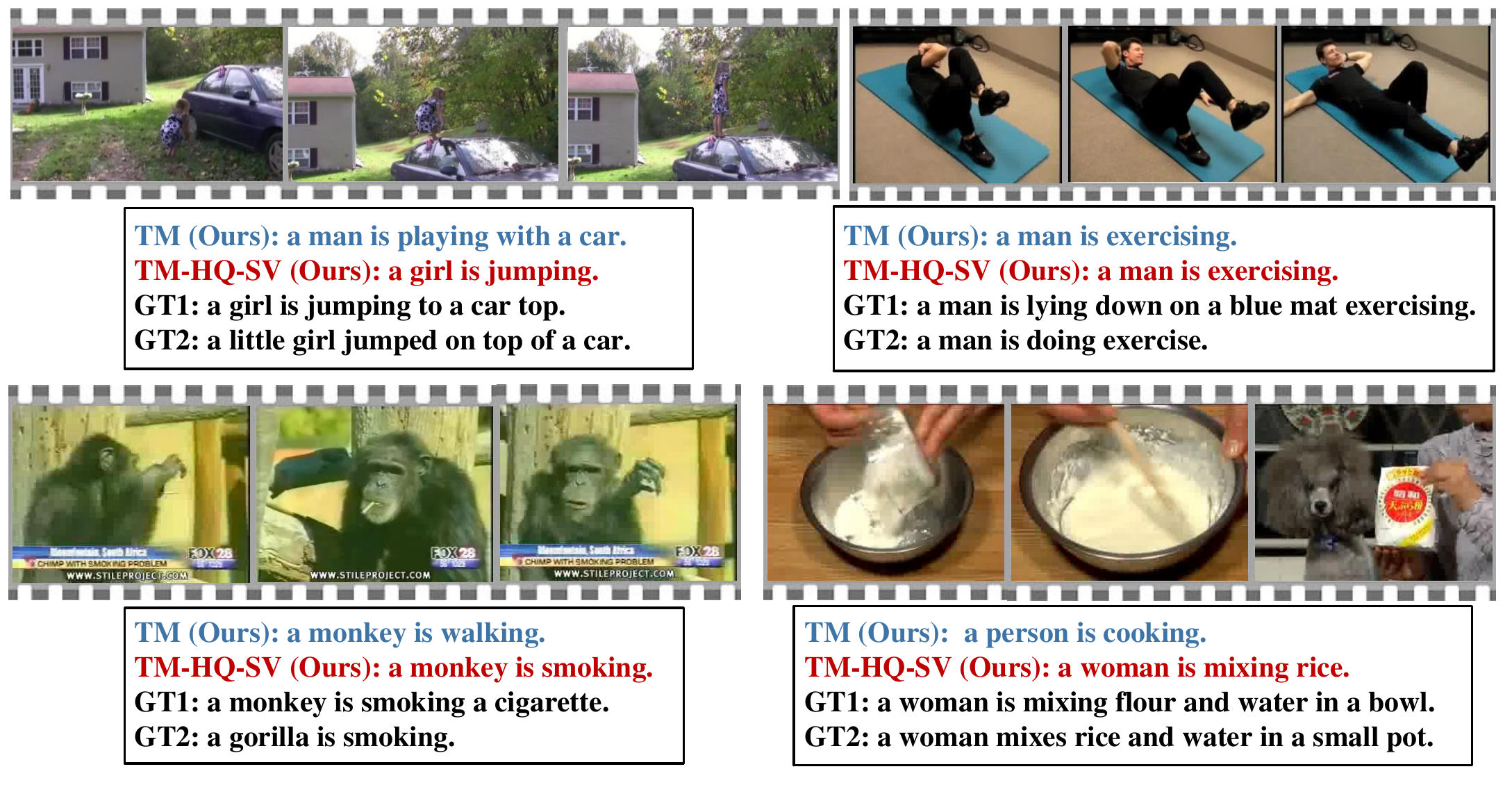}\\
        \caption{Examples of generated captions on MSVD. GT1 and GT2 are ground truth captions.}
        \label{fig:captions}
    \end{figure*}

\section{Experimental Results}
\label{sec:experiments}

    \subsection{Datasets}
        \noindent \textbf{MSVD}: We evaluate our video captioning models on the Microsoft Research Video Description Corpus \cite{Chen2011Collecting}. MSVD consists of 1,970 video clips typically depicting a single activity, downloaded from YouTube. Each video clip is annotated with multiple human generated descriptions in several languages. We only use the English descriptions, about 41 descriptions per video. In total, the dataset consists of 80,839 video/description pairs. Each description on average contains about 8 words. We use 1,200 videos for training, 100 videos for validation and 670 videos for testing, as provided by previous work \cite{Guadarrama2013YouTube2Text}.

        \pheadB{MSR-VTT}: We also evaluate on the MSR Video-to-Text (MSR-VTT) dataset \cite{Xu2016MSR}, a new large-scale video benchmark for video captioning. MSR-VTT provides 10,000 web video clips. Each video is annotated with about 20 natural sentences. Thus, we have 200,000 video-caption pairs in total. Our video captioning models are trained and hyper-parameters are selected using the official training and validation set, which consists of 6,513 and 497 video clips respectively. And models are evaluated using the test set of 2,990 video clips.

    \subsection{Preprocessing}
        \noindent \textbf{Visual Features}: We extract two kinds of visual features, temporal features and motion features. We use a pretrained ResNet-152 model \cite{He2015Deep} to extract temporal features, obtaining one fixed-length feature vector for every frame. We use a pretrained C3D \cite{Du2014C3D} to extract motion features. The C3D net reads in a video and emits a fixed-length feature vector every 16 frames.

        \noindent \textbf{Semantic Attributes}: While MSVD and MSR-VTT are standard video caption datasets, they do not come with tags, titles, or other semantic information about the videos. Nevertheless, we can reproduce a setting with semantic attributes by extracting attributes from captions. First, we invoke the Stanford Parser \cite{Klein2003Accurate} to parse captions and choose the \emph{nsubj} edges to find the subject-verb pairs for each caption. We then select the most frequent subject and verb across captions of each video as the high-quality semantic attributes.
        These attributes can be used to evaluate our models under a high-quality attribute condition. Next, we can use the high-quality semantic attributes of the training set to train a model to predict semantic attributes for the test set. Such attributes are used to evaluate our model under low-quality semantic attribute conditions.

        For our experiments, we consider three models to predict semantic attributes. The first one (NN) is to perform a nearest-neighbor search on every frame of the training set to retrieve similar ones for every frame of each test video based on ResNet-152 features and select the most frequent attributes. The second one (SVM) is to train SVMs for the top 100 frequent attributes in the training set and predict semantic attributes for test videos based on a mean pooling of ResNet-152 features. The third one (HRNE) is to train two hierarchical recurrent neural encoders \cite{Pan2015Hierarchical} to predict the subject and verb separately based on temporal ResNet-152 features.

    \subsection{Evaluation Metrics}
        We rely on four standard metrics, BLEU \cite{Papineni2002.BLEU}, METEOR \cite{Banerjee2005METEOR}, CIDEr \cite{Vedantam2015CIDEr} and ROUGE-L \cite{Lin2004ROUGE} to evaluate our methods. These are commonly used in image and video captioning tasks, and allow us to compare our results against previous work. We use the Microsoft COCO evaluation server \cite{Chen2015Microsoft}, which is widely used in previous work, to compute the metric scores. Across all three metrics, higher scores indicate that the generated captions are assessed as being closer to captions created by humans.

    \subsection{Experimental Settings}
        The number of hidden units in the input multimodal layer and in the LSTM are both $512$. The activation function of the LSTM is $\tanh$ and the activation functions of both multimodal layers are $\linear$.
        The dropout rates of both the input and output multimodal layers are set to $0.5$.
        We use pretrained 300-dimensional GloVe \cite{Pennington2014Glove} vectors as our word embedding matrix. We rely on the RMSPROP algorithm  \cite{Tieleman2012rmsprop} to update parameters for better convergence, with the learning rate $10^{-4}$.
        The beam size during sentence generation is set to 5. Our system is implemented using the Theano \cite{Bastien2012Theano,J2010Theano} framework.

    \subsection{Results}

        \noindent \textbf{Visual only}: First, for comparison, we show the result of only using visual attention at first. Specifically, we only use the temporal features and motion features (TM), by removing the semantic branch with other components of our model unchanged. To evaluate the effectiveness of different sorts of visual cues, we also report the results of using only temporal features (T) and using only motion features (M). We compare our methods with six state-of-the-art methods: LSTM-YT \cite{Venugopalan2015Translating}, S2VT \cite{Venugopalan2015Sequence}, TA \cite{Yao2015Describing}, LSTM-E \cite{Pan2016Jointly}, HRNE-A \cite{Pan2015Hierarchical}, and h-RNN \cite{Yu2015Video}. Table \ref{tab:results1} provides a comparison of these systems on the MSVD dataset. Since some of the previous work uses different features, we also run experiments for some of them whose source code are provided by the authors, or we re-implement the models described in their papers, and then evaluate them using our features. The corresponding extra results are marked by `$\ast$'.

        We observe that even just with temporal features alone, we obtain fairly good results, which implies that the attention model in our approach is useful. Combining temporal and motion features, we see that our method can outperform previous work, confirming that our attention model with multimodel layers can extract useful information from temporal and motion features effectively. In fact, the TA, LSTM-E studies also employ both temporal and motion features, but do not have a separate motion attention mechanism. And the h-RNN study only considers attention after the sentence generator. Instead our attention mechanism operates both before and after the sentence generator, enabling it to attend to different aspects during the analysis and synthesis processes for a single sentence. The results on the MSR-VTT dataset are shown in Table \ref{tab:results3}. They are consistent in that they also show that the combined attention for temporal and motion features obtains better results.

        \pheadB{Multi-Faceted Attention}: To show the influence of our multi-faceted attention with additional semantic cues, we first consider the low-quality semantic attributes. Tables \ref{tab:results2} and \ref{tab:results3} provide results using low-quality attributes obtained via our NN (TM-P-NN), SVM (TM-P-SVM), and HRNE (TM-P-HRNE) methods described above. We find that the results for NN and SVM are sometimes slightly worse than only using visual attention, which means that too low-quality attributes do not help in improving the quality.
        It appears that these methods are rather unreliable and introduce significant noise. HRNE fares slightly better than using only visual attention, as it combines top-down and bottom-up approaches to obtain more stable and reliable results.

        Then, we consider the high-quality semantic attributes, subject and verb (TM-HQ-SV), derived from the ground truth captions. We also report the performance of only using the subject (TM-HQ-S) or the verb (TM-HQ-V) individually. These results, too, are included in Tables \ref{tab:results2} and \ref{tab:results3}. We find that our method is able to exploit high-quality subject and verb attributes to outperform other methods by very large margins. Even using just a single semantic attribute yields very strong results. Here, verb information proves slightly more informative than the subject, indicating that identifying actions in videos remains more challenging than identifying important objects in a video.

        Overall, we observe that our method, with just two high-quality features, approaches human-level scores in terms of the METEOR and CIDEr metrics.
        For this, we randomly selected one caption for each video in the test set and evaluate this caption by removing them from the ground truth. Although not perfect, such results (Human) can be viewed as an estimation of the human-level performance. The BLEU scores of our method are in fact even greater than the human-level ones, since humans often prefer generating longer captions, which tend to obtain lower BLEU scores. Several studies, including on caption generation, have concluded that BLEU is not a sufficiently reliable metric in terms of replicating human judgment scores \cite{Kulkarni2013BabyTalk,Vedantam2015CIDEr}. Figure \ref{fig:captions} shows several example captions generated by our approach for MSVD videos.

        To further investigate the influence of noise, we randomly select genuine high-quality subject and verb attributes and replace them with random incorrect ones. Figure \ref{fig:result} provides the results on MSVD. These result show that even when adding $50\%$ noise, the results are better than just using regular visual attention. With extremely strong noise levels, the results are worse than only using visual attention, but are still maintained at a certain level. This shows that we are likely to benefit from further semantic attributes such as tags, titles, comments, and so on, which are often available for online videos, even if they are noisy.

\section{Conclusion}
        We have proposed a novel method for video captioning based on an extensible multi-faceted attention mechanism, outperforming previous work by large margins.

        Even without semantic attributes, our method outperforms state-of-the-art approaches using visual features.
        With just two high-quality semantic attributes, the results become competitive with human results across a range of metrics.
        This opens up important new avenues for future work on exploring the large space of potential additional forms of semantic cues and attributes.

\section{Acknowledgments}
        This work was supported in part by the National Basic Research Program of China Grant 2011CBA00300, 2011CBA00301 and the National Natural Science Foundation of China Grant 61033001, 61361136003.

\bibliographystyle{acl2012}

\end{document}